\documentclass[letterpaper]{article} 
\usepackage{aaai2026}  
\usepackage{times}  
\usepackage{helvet}  
\usepackage{courier}  
\usepackage[hyphens]{url}  
\usepackage{graphicx} 
\usepackage{natbib}  
\usepackage{caption} 
\usepackage{algorithm}
\usepackage{algorithmic}
\usepackage{amsmath}
\usepackage{amsfonts}
\usepackage{booktabs}
\usepackage{newfloat}
\usepackage{listings}
\DeclareCaptionStyle{ruled}{labelfont=normalfont,labelsep=colon,strut=off} 
\floatstyle{ruled}
\newfloat{listing}{tb}{lst}{}
\floatname{listing}{Listing}
\title{Rethinking Multimodal Point Cloud Completion: \\ A Completion-by-Correction Perspective}
\author{
    Wang Luo,
    Di Wu\thanks{Corresponding author.},
    Hengyuan Na,
    Yinlin Zhu,
    Miao Hu,
    Guocong Quan
}
\affiliations{
    Sun Yat-sen University, Guangzhou, China\\

    Guangzhou Yunshan Research Institute of Artificial Intelligence Security, Guangzhou, China\\
    
    luow69@mail2.sysu.edu.cn, wudi27@mail.sysu.edu.cn, neihy@mail2.sysu.edu.cn, zhuylin27@mail2.sysu.edu.cn,
    humiao5@mail.sysu.edu.cn, quangc@mail.sysu.edu.cn
}

\begin{document}

\maketitle

\begin{abstract}
Point cloud completion aims to reconstruct complete 3D shapes from partial observations, which is a challenging problem due to severe occlusions and missing geometry. Despite recent advances in multimodal techniques that leverage complementary RGB images to compensate for missing geometry, most methods still follow a \textit{Completion-by-Inpainting} paradigm, synthesizing missing structures from fused latent features. We empirically show that this paradigm often results in structural inconsistencies and topological artifacts due to limited geometric and semantic constraints. To address this, we rethink the task and propose a more robust paradigm, termed \textit{Completion-by-Correction}, which begins with a topologically complete shape prior generated by a pretrained image-to-3D model and performs feature-space correction to align it with the partial observation. This paradigm shifts completion from unconstrained synthesis to guided refinement, enabling structurally consistent and observation-aligned reconstruction. Building upon this paradigm, we introduce PGNet, a multi-stage framework that conducts dual-feature encoding to ground the generative prior, synthesizes a coarse yet structurally aligned scaffold, and progressively refines geometric details via hierarchical correction. Experiments on the ShapeNetViPC dataset demonstrate the superiority of PGNet over state-of-the-art baselines in terms of average Chamfer Distance (-23.5\%) and F-score (+7.1\%).
\end{abstract}

\begin{links}
    \link{Code}{https://github.com/RobWonn/PGNet}
\end{links}

\section{Introduction}

With the popularity of LiDAR and RGB-D cameras, point clouds have emerged as a fundamental 3D representation and are widely adopted in diverse AI applications, including autonomous driving \cite{chen20203d}, augmented reality \cite{wang2023pointshopar}, and robotics \cite{varley2017shape}. However, point clouds captured by these sensors are often sparse and incomplete due to occlusion, light reflection, and limited resolution, hindering the performance of downstream tasks \cite{liang2019multi, nie2021rfd}. Consequently, point cloud completion, which aims to recover complete point clouds from partial input, has become an indispensable task.

\begin{figure}[t]
\centering
\includegraphics[width=1.0\columnwidth]{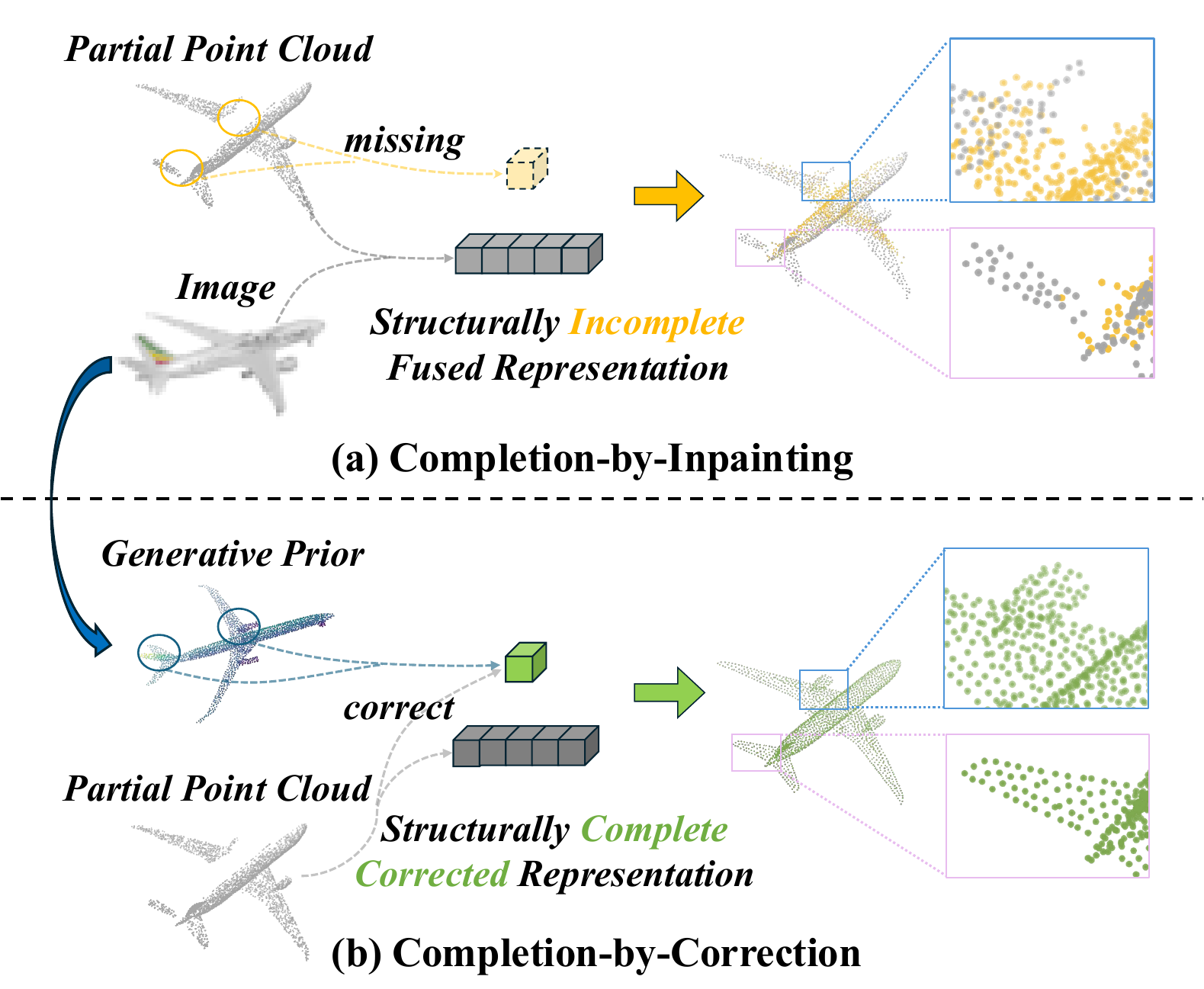} 
\caption{Comparison of two point cloud completion paradigms. (a) Completion-by-Inpainting synthesizes missing geometry from incomplete representation, often introducing artifacts. (b) Our Completion-by-Correction leverages a complete generative prior, correcting it by grounding to the partial observation for more consistent structures.}
\label{plot-intro}
\end{figure}

Traditional methods \cite{kazhdan2013screened, pauly2008discovering, mitra2013symmetry, berger2014state} for point cloud completion relied on geometric heuristics or template matching, constructing complete shapes by aligning geometric rules or retrieving similar instances from predefined databases. In recent years, deep learning-based unimodal approaches \cite{yuan2018pcn, yu2021pointr, cai2024orthogonal} have made significant strides by learning shape priors from large-scale datasets and directly synthesizing completions from partial inputs. However, when relying only on geometric input without other contextual cues, it remains difficult to judge whether missing parts stem from occlusion or represent actual voids in the object's structure, which often leads to sub-optimal completion performance.

Motivated by the human capability to perceive 3D structures from 2D views, recent studies explore multimodal learning for point cloud completion, using RGB images as a readily accessible source of additional texture and semantic information. For example, CSDN \cite{zhu2023csdn} treats completion as a style transfer task by injecting image features into a folding-based decoder, and introduces a dual-refinement module leveraging global image guidance and local geometric cues. XMFNet \cite{aiello2022cross} uses stacked cross- and self-attention layers to fuse image and point cloud features in the latent space, reconstructing shapes via multiple prediction branches. In contrast, EGIInet \cite{xu2024explicitly} adopts an explicitly guided interaction strategy with a novel loss to align structural information across modalities before fusion.

Despite their effectiveness, these methods face \textbf{Critical Limitations}: As illustrated in Figure~\ref{plot-intro}(a), most existing approaches adopt a {\textit{Completion-by-Inpainting}} paradigm, synthesizing missing geometry from fused visual and geometric features. We empirically demonstrate that this process is inherently uncertain and becomes unreliable under severe degradation, as the network operates without an explicit structural scaffold and must hallucinate structure from limited guidance (Sec.~\ref{sec: exp q1}). As a result, although the generated completions appear semantically plausible, they frequently exhibit structural inconsistencies and topological artifacts.

To this end, we rethink the task of multimodal point cloud completion and propose a more robust paradigm, termed the {\textit{Completion-by-Correction}}. As illustrated in Figure~\ref{plot-intro}(b), rather than synthesizing geometry from an incomplete fused representation, we begin with a topologically complete and semantically meaningful shape prior generated by a pretrained image-to-3D model. We subsequently perform feature-space correction to align this prior with the partial observation, instead of relying on direct geometric fusion \cite{zhang2021view, wei2025pcdreamer}, which is often undermined by pose and scale misalignment introduced by model bias and image ambiguity. This paradigm shifts the task from unconstrained synthesis to guided refinement over an observation-consistent representation, making it more well-posed and robust to geometric uncertainty.

Building upon this insight, we introduce \textbf{P}rior\textbf{G}round\textbf{Net} (PGNet), a novel framework that realizes the Completion-by-Correction paradigm through a carefully designed three-stage process. (1) {\textit{Corrective Dual-Feature Encoding}}, the generative prior is corrected by grounding its features in the partial observation via parallel encoding and semantic correspondence; (2) {\textit{Grounded Seed Generation}}, synthesizes a coarse but topologically complete point cloud that serves as a structural scaffold aligned with the observation; (3) {\textit{Hierarchical Grounded Refinement}}, iteratively refines the coarse scaffold by aggregating dual-source features for each point, capturing high-fidelity geometry from the observation and structural context from the prior, predicting displacements informed by shape context.

\textbf{Our Contributions}: (1) \textbf{Paradigm Identification.} We introduce the Completion-by-Correction paradigm, which reformulates multimodal point cloud completion. Instead of inpainting missing regions, our paradigm corrects a structurally complete generative shape prior by grounding it in the partial observation, reducing geometric ambiguity and structural artifacts. (2) \textbf{Novel Framework.} We propose PGNet, a framework that implements our paradigm through three dedicated stages: Corrective Dual-Feature Encoding, Grounded Seed Generation, and Hierarchical Grounded Refinement. (3) \textbf{SOTA Performance.} PGNet consistently achieves state-of-the-art performance on the ShapeNetViPC dataset, reducing average Chamfer Distance by 23.5\% and improving average F-score by 7.1\% over prior methods.

\begin{figure*}[t]
\centering
\includegraphics[width=0.9\textwidth]{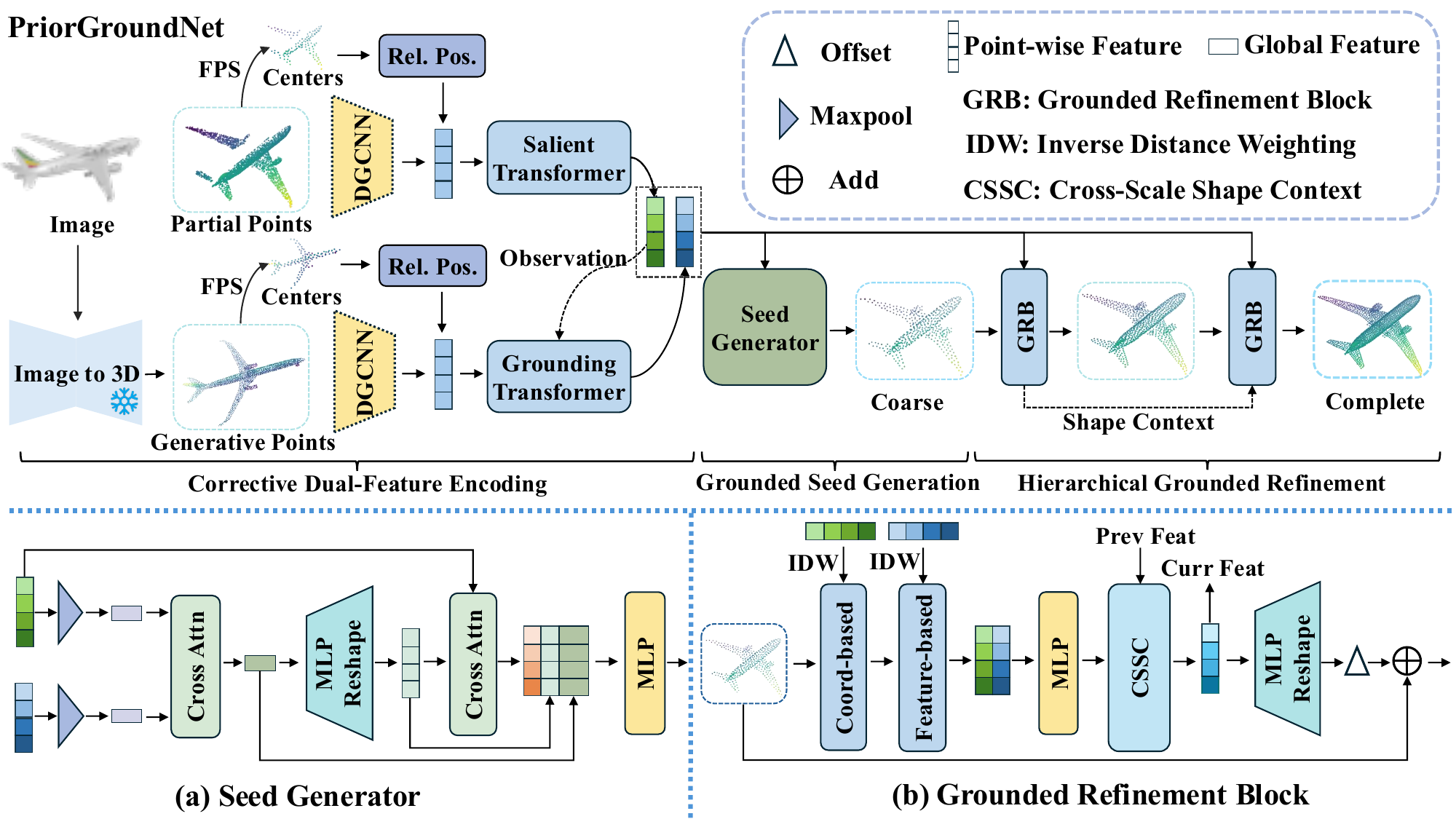} 
\caption{The overall architecture of PriorGroundNet (PGNet), which follows the Completion-by-Correction paradigm and consists of three stages: Corrective Dual-Feature Encoding, Grounded Seed Generation, and Hierarchical Grounded Refinement. (a) The detailed structure of the Seed Generator module, which produces a coarse but complete point cloud by grounding semantic seeds. (b) The architecture of the Grounded Refinement Block (GRB), which hierarchically enhances geometric detail using dual-source feature.}
\label{overview}
\end{figure*}

\section{Related Works}
\subsection{Unimodal Point Cloud Completion}
Traditional methods include geometry-based and alignment-based techniques. Geometry-based methods \cite{berger2014state, davis2002filling, nealen2006laplacian, sarkar2017learning} infer missing structures using geometric priors such as surface smoothness or local symmetry. These approaches typically interpolate or replicate geometric patterns but often struggle with incomplete or complex structures. Alignment-based \cite{mitra2006partial, mitra2013symmetry, pauly2008discovering, sipiran2014approximate, sung2015data} methods retrieve and fit models from shape databases via shape matching, part assembly, or template deformation. However, their applicability is constrained by computational cost, noise sensitivity, and limited dataset coverage.

Deep learning methods have demonstrated superior performance by learning data-driven shape priors. PCN \cite{yuan2018pcn} introduced a PointNet-based encoder and a FoldingNet-style \cite{yang2018foldingnet} decoder, enabling end-to-end completion without strong geometric assumptions. TopNet \cite{tchapmi2019topnet} adopted a tree-structured decoder to support hierarchical generation, while PF-Net \cite{huang2020pf} enhanced feature fusion and cascaded refinement. Transformer-based architectures further advanced this line of research: PoinTr \cite{yu2021pointr} reformulated completion as set-to-set translation and incorporated a geometry-aware module; SnowFlakeNet \cite{xiang2021snowflakenet} applied iterative snowflake-like deconvolution. SeedFormer \cite{zhou2022seedformer} proposed a shape representation using patch seeds that capture both global structure and local details. PointAttN \cite{wang2024pointattn} employs attention mechanisms to capture local and global structures among unordered points without explicit region partitioning. CRA-PCN \cite{rong2024cra} and PointCFormer \cite{zhong2025pointcformer} improved local geometric modeling through cross-resolution attention and relation-based metrics. Recent works \cite{wei2025pcdreamer, yan2025symmcompletion, kasten2023point, chu2025digging} also explored diffusion models \cite{ho2020denoising}, symmetry priors, and context-aware refinement strategies.

\subsection{Multimodal Point Cloud Completion}
Multimodal methods leverage auxiliary modalities, typically a single RGB image, to guide point cloud completion. Early works such as ViPC \cite{zhang2021view} concatenated coarse image-derived point clouds with partial inputs for structural enhancement. More recent methods employ advanced fusion strategies. XMFNet \cite{aiello2022cross} performs implicit fusion using layered attention and reconstructs shapes via multi-branch prediction. CSDN \cite{zhu2023csdn} treats completion as a style transfer task by embedding image features into the decoding process and refining output through joint image and geometry cues. EGIInet \cite{xu2024explicitly} enhances structural understanding through explicitly guided interaction and modality alignment. CDPNet \cite{du2024cdpnet} adopts a patch-based strategy to densify coarse point clouds and extracts a cross-modal style code to guide structural detail generation. DMF-Net \cite{mao2025dmf} introduces a dual-channel fusion framework with a shape-aware upsampling transformer, enabling balanced exploitation of image and point cloud features.

In contrast to prior works that synthesize missing geometry directly from fused features, we reformulate completion as a correction process over a complete shape prior. By grounding this prior in partial observations, our approach transforms the task from uncertain generation to observation-guided structural refinement.

\section{Methodology}

\subsection{Overview}
The overall architecture of PGNet is shown in Figure~\ref{overview}, which consists of three stages: Corrective Dual-Feature Encoding, Grounded Seed Generation, and Hierarchical Grounded Refinement. We will detail each module in the following.

\subsubsection{Problem Definition.}
Given a partial point cloud observation $P_o \in \mathbb{R}^{M \times 3}$ and its corresponding single-view RGB image $I \in \mathbb{R}^{H \times W \times 3}$, the objective of multimodal point cloud completion is to reconstruct a complete and high-fidelity point cloud $P_{gt} \in \mathbb{R}^{N \times 3}$. Here, $M$ and $N$ denote the number of points in the partial and ground truth point clouds, respectively.

\subsubsection{The Completion-by-Correction Paradigm.}
More robust and well-posed, the Completion-by-Correction paradigm replaces ill-posed synthesis with the correction of a generative shape prior, guided by partial observations. It initializes from a topologically complete and semantically meaningful shape prior $P_g = \mathcal{G}(I)$, generated by a pre-trained image-to-3D model $\mathcal{G}$. While $P_g \in \mathbb{R}^{N_g \times 3}$ provides a strong structural scaffold, it may contain geometric inaccuracies due to model bias or input ambiguity. The goal is thus reframed as learning a correction function $\mathcal{T}$ that grounds $P_g$ in the partial observation $P_o$, yielding a final output $P = \mathcal{T}(P_g, P_o)$ that aligns with the ground truth $P_{gt}$.

\subsection{Corrective Dual-Feature Encoding}
A key challenge in this paradigm is the discrepancy between the generative prior $P_g$ and the partial observation $P_o$, which often differ in scale, pose, and point distribution and may contain hallucinated geometry. We therefore encode $P_g$ and $P_o$ in parallel and ground the prior features on those from the partial observation in feature space, enabling robust alignment and providing a corrected context for subsequent generation.

\subsubsection{Partial Point Cloud Encoder.}
The partial point cloud $P_o$ provides reliable geometric evidence. As illustrated in Figure~\ref{overview}(a), we adopt a hierarchical local feature aggregation strategy following PoinTr~\cite{yu2021pointr}: Farthest Point Sampling (FPS) selects $N_e$ representative points, and a lightweight DGCNN~\cite{wang2019dynamic} aggregates local features around each point.
This yields point centers $C_o = \{c_{o,i}\}_{i=1}^{N_e}$ and an initial feature tensor $F'_o \in \mathbb{R}^{N_e \times D}$.

To alleviate pose and scale discrepancies between $P_o$ and $P_g$, we use a learnable relative position embedding $\Phi$ that encodes the local spatial layout of each patch and is added to the initial features:
\begin{equation}
    F''_o = F'_o + \Phi(C_o).
\end{equation}

While DGCNN effectively captures local geometric structure, it is limited in modeling long-range dependencies and handling non-uniform point densities. We therefore introduce the Salient Transformer, a dual-branch architecture that integrates global and local context. The global branch applies multi-head self-attention (MHSA) to $F''_o$ to produce long-range context features $A_o$, and the local branch aggregates $k$-nearest-neighbor patterns $X_o$ via shared MLPs and max pooling. To adaptively combine the two sources of information, a learnable salience gate dynamically fuses global and local features:
\begin{equation}
\begin{split}
    G_o &= \sigma(\text{MLP}([A_o, X_o])), \\
    F_o &= (1 - G_o) \odot A_o + G_o \odot X_o,
\end{split}
\end{equation}
where $\sigma$ denotes the sigmoid function and $\odot$ denotes element-wise multiplication. This mechanism allows the network to emphasize global context in sparse regions and local detail where fine geometry is critical.

\subsubsection{Generative Point Cloud Encoder.}
We apply the same hierarchical encoding to $P_g$, obtaining representative centers $C_g$ and initial features $F'_g \in \mathbb{R}^{N_e \times D}$. Adding the shared positional encoding yields
\begin{equation}
    F''_g = F'_g + \Phi(C_g).
\end{equation}
On top of $F''_g$, we build a Grounding Transformer that refines the generative prior in feature space under the guidance of the partial observation. It processes $F''_g$ through two parallel branches: a self-attention branch that applies MHSA to produce contextual features $A_g$ encoding internal structural priors of $P_g$, and a grounding branch that performs cross-attention with $F''_g$ as queries and $F_o$ as keys and values to obtain observation-aligned features $X_g$. To adaptively balance these complementary cues, we reuse the same salience gate formulation as in the Salient Transformer:
\begin{equation}
\begin{split}
    G_g &= \sigma(\text{MLP}([A_g, X_g])), \\
    F_g &= (1 - G_g) \odot A_g + G_g \odot X_g,
\end{split}
\end{equation}
The resulting features $F_g = \{f_{g,i}\}_{i=1}^{N_e}$ form a corrected, observation-aware representation of the generative prior, which serves as the foundation for subsequent grounded seed generation and hierarchical refinement. Conceptually, the Salient Transformer enhances the reliability of $F_o$ by balancing global and local evidence, while the Grounding Transformer injects this reliable observation signal into the generative prior.

\subsection{Grounded Seed Generation}
The goal of this stage is to synthesize a structurally complete yet geometrically grounded scaffold as the basis for subsequent refinement. Consistent with our Completion-by-Correction paradigm, we leverage the topological completeness of the generative prior and align it with the geometric fidelity of the partial observation.

As illustrated in Figure~\ref{overview}(a), we first perform max pooling on $F_g$ and $F_o$ to extract their global representations $\hat{F}_g$ and $\hat{F}_o$. A cross-attention module is then applied, with $\hat{F}_o$ as the query and $\hat{F}_o$ as key and value, yielding a fused global feature $\hat{F}_{\text{fused}}$ that encodes an observation-aware yet topologically complete shape representation.

Next, instead of directly regressing point coordinates from $\hat{F}_{\text{fused}}$, we generate a structured set of $N_c$ seed features $F_{\text{seed}} \in \mathbb{R}^{N_c \times D}$ by projecting $\hat{F}_{\text{fused}}$ through an MLP and reshaping the output. This operation, inspired by the PixelShuffle~\cite{shi2016real} mechanism, effectively expands the global feature into a spatially organized seed representation, enabling the network to model inter-point dependencies and produce coherent, structurally consistent layouts.

To further incorporate geometric grounding, we apply cross-attention with $F_{\text{seed}}$ as the query and $F_o$ as the key and value, producing grounded features $F_{\text{gr}}$. The final coarse point cloud $P_c$ is then obtained by feeding the concatenation of the replicated global feature, seed features, and grounded features into an MLP:
\begin{equation}
P_c = \text{MLP}([\text{Replicate}(\hat{F}_{\text{fused}}, N_c), F_{\text{seed}}, F_{\text{gr}}]).
\end{equation}

In this way, global priors propose a complete seed layout, while point-wise grounding from $F_o$ adjusts it towards the observed geometry, yielding a structurally complete and observation-aware scaffold for later refinement.

\subsection{Hierarchical Grounded Refinement}
Unlike inpainting-based methods that decode abstract features into dense point sets, our approach refines geometry using corrected features from feature-space grounding. The coarse output $P_c$ offers a complete but low-resolution scaffold aligned with the observation. Built on this, we introduce a hierarchical architecture of K stacked Grounded Refinement Blocks (GRBs) (Figure~\ref{overview}(b)), which progressively improve geometric fidelity via localized matching guided by both the observation and corrected prior.

Each GRB contains two components: (1) Dual-Source Feature Association, which retrieves semantically aligned local patterns from both sources; and (2) Structure-Aware Upsampling, which captures shape context and predicts localized displacements to refine spatial layout.

\subsubsection{Dual-Source Feature Association.}
Given an input point set $P_{\text{in}} \in \mathbb{R}^{N_{\text{in}} \times 3}$ from the previous stage, we enrich each point $p_i$ with localized features by querying both the partial observation and the corrected generative prior. This association propagates observation guidance while injecting structurally plausible patterns from the prior.

We begin by querying features from the partial observation. Given the center points $C_o$ and their associated features $F_o$, we employ inverse distance weighting (IDW)~\cite{qi2017pointnet++} to interpolate features for each query point $p_i$ based on its $k$ nearest neighbors in $C_o$:
\begin{equation}
\begin{split}
f_{\text{interp}, o}(p_i) &= \frac{\sum_{j \in \mathcal{N}_k(p_i, C_o)} w_j f_{o,j}}{\sum_{j \in \mathcal{N}_k(p_i, C_o)} w_j}, \\
w_j &= \frac{1}{\|p_i - c_{o,j}\|_2},
\end{split}
\end{equation}
where $\mathcal{N}_k(p_i, C_o)$ denotes the $k$ nearest neighbors of $p_i$ in Euclidean space. To integrate prior structure, we additionally query from $F_g$. As direct spatial interpolation may fail due to misalignment between $P_o$ and $P_g$, we instead interpolate in the feature space. Specifically, each $f_{\text{interp}, o}(p_i)$ is used to find its $k$ nearest neighbors in $F_g$, and IDW is applied based on feature distance:
\begin{equation}
\begin{split}
f_{interp, g}(p_i) &= \frac{\sum_{j \in \mathcal{N}_k(f_{interp, o}(p_i), F_g)} w'_j f_{g,j}}{\sum_{j \in \mathcal{N}_k(f_{interp, o}(p_i), F_g)} w'_j}, \\
w'_j& = \frac{1}{\|f_{interp, o}(p_i) - f_{g,j}\|_2}.
\end{split}
\end{equation}
The final dual-source representation is formed by concatenating both:
\begin{equation}
    f_{as}(p_i) = [f_{interp, o}(p_i), f_{interp, g}(p_i)].
\end{equation}

This dual-source association is critical for resolving coordinate misalignment and grounding geometric features in partial observations. Structural patterns from the generative prior are selectively integrated under observation guidance, ensuring that only semantically consistent regions contribute to refinement. 

\subsubsection{Structure-Aware Upsampling.}
Given the fused dual-source features for each point in $P_{\text{in}}$, we aim to refine geometry by enhancing structural fidelity and increasing point density. To this end, we aggregate multi-scale shape context so that each point receives geometric information from its local neighborhood in the previous resolution, enabling more informed prediction of point-wise displacements.

Central to this stage is the Cross-Scale Shape Context (CSSC) module, which enhances each point in the current resolution by attending to geometrically relevant features from a source point set. This source consists of spatial coordinates $P_k$ and their associated features, derived from the contextual outputs $F_{ctx}^{prev}$ of the previous refinement stage.
Note that in the first block, where no prior context is available, both $P_k$ and the associated features default to the input point set $P_{\text{in}}$ and its projected features. To support cross-scale attention, the fused dual-source features $F_{as}$ are projected to obtain a set of query features $F_q$. For each query point $p_i \in P_{\text{in}}$, we identify its $k$ nearest neighbors in $P_k$, and project the corresponding features into key and value vectors $k_j$ and $v_j$, while $f_{q,i}$ is projected into a query vector $q_i$. Following geometric transformer design~\cite{zhao2021point}, attention weights are computed based on both feature similarity and relative spatial position:
\begin{equation}
    \alpha_{ij} = \frac{\exp(\text{MLP}(q_i - k_j + \Phi(p_i - p_{k,j})))}{\sum_{l \in \mathcal{N}_k(p_i, P_k)} \exp(\text{MLP}(q_i - k_l + \Phi(p_i - p_{k,l})))}
\end{equation}
where $\Phi$ denotes the relative positional encoding. The final contextual feature $f_{ctx,i}$ is computed by weighted aggregation of neighbor values, with a residual connection:
\begin{equation}
    f_{ctx,i} = f_{q,i} + \sum_{j \in \mathcal{N}_k(p_i)} \alpha_{ij} v_j.
\end{equation}
This CSSC module enriches each point with geometric structure from the previous resolution, facilitating spatially consistent upsampling. Once the contextual features $F_{ctx} = \{f_{ctx,i}\}_{i=1}^{N_{\text{in}}}$  are computed for all input points, they are used to predict displacements for upsampling. An MLP first expands the channel dimension of $F_{ctx}$ and the result is reshaped to generate $r$ distinct displacement vectors for each point, which can be formulated as
\begin{equation}
    \Delta = \text{Reshape}(\text{MLP}(F_{ctx})),
\end{equation}
where $\Delta \in \mathbb{R}^{r N_{\text{in}} \times 3}$ represents the complete set of predicted offsets and $r$ is the upsampling rate. The final upsampled point cloud $P_{up}$ is obtained by replicating the input point set and adding the predicted displacements:
\begin{equation}
P_{up} = \text{Replicate}(P_{\text{in}}, r) + \Delta.
\end{equation}

\subsection{Loss Function}
Following prior work~\cite{aiello2022cross, mao2025dmf}, we adopt the L1 Chamfer Distance (L1-CD) as the training objective. Given a predicted point cloud $P_{pred}$ and the ground truth $P_{gt}$, the L1-CD is defined as:
\begin{equation}
\begin{split}
    \mathcal{L}_{\text{CD}}(P_{pred}, P_{gt}) ={}& \frac{1}{|P_{pred}|} \sum_{x \in P_{pred}} \min_{y \in P_{gt}} |x - y|_1 \\
                                             & + \frac{1}{|P_{gt}|} \sum_{y \in P_{gt}} \min_{x \in P_{pred}} |y - x|_1,
\end{split}
\end{equation}
where $|\cdot|_1$ denotes the L1 norm. The first term encourages predicted points to align closely with the ground truth, while the second term ensures complete coverage of the ground truth by the prediction.

To promote consistency across refinement stages, we directly supervise the coarse output $P_c$ and each upsampled point cloud $\{P^{(k)}\}_{k=1}^K$ using the high-resolution ground truth $P_{gt}$. This encourages intermediate predictions to approximate the target distribution early on, facilitating convergence and improving geometric fidelity throughout the hierarchy. The final loss is computed as the unweighted average of the L1-CD values across all prediction levels:
\begin{equation}
\mathcal{L} = \frac{1}{K+1} \left( \mathcal{L}_{\text{CD}}(P_c, P_{gt}) + \sum_{k=1}^{K} \mathcal{L}_{\text{CD}}(P^{(k)}, P_{gt}) \right).
\end{equation}

\section{Experiments}
In this section, we present a comprehensive evaluation of PGNet. We first describe the dataset and implementation details (Sec.4.1). Then, we aim to address the following questions: \textbf{Q1}: How does PGNet compare with state-of-the-art methods (Sec.4.2)? \textbf{Q2}: How does the proposed Completion-by-Correction paradigm compare with Completion-by-Inpainting strategies (Sec.4.3)? \textbf{Q3}: How important are individual components of PGNet to its overall performance (Sec.4.4)? Moreover, we provide further experimental investigations about the generalization performance (\textbf{Q4}) and the robustness to variations in the generative prior (\textbf{Q5}) of PGNet in the supplementary material.


\subsection{Dataset and Implementation Details}
\subsubsection{Dataset.}
Following prior works \cite{aiello2022cross, xu2024explicitly}, we train and evaluate on the ShapeNet-ViPC~\cite{zhang2021view} dataset, comprising 38,328 objects across 13 categories. Each sample consists of a partial input ($M=2048$ points), its corresponding image, and a ground truth completion ($N=2048$ points).

\subsubsection{Implementation Details.}
The network is trained end-to-end in PyTorch \cite{paszke2019pytorch} using the AdamW optimizer with an initial learning rate of $2 \times 10^{-4}$ and a cosine annealing schedule. Each of the eight categories is trained separately for 100,000 iterations with a batch size of 192 on NVIDIA RTX 4090 GPU. To generate priors, we use the Trellis \cite{xiang2025structured} image-to-3D model to predict a mesh from each input image and apply Poisson disk sampling on the mesh surface to extract 2048 points.

The network architecture adopts DGCNN-based encoders that extract $N_e = 128$ local feature centers. The Seed Generator produces a 512-point coarse scaffold. The encoding modules apply 6-head attention with 768-dimensional hidden layers for semantic representation learning, while other attention modules use 4 heads and 256 dimensions. All non-attention layers use $D = 256$ feature dimensions, and $k = 8$ for feature interpolation during refinement. The hierarchical refinement stage includes $K = 2$ Grounded Refinement Blocks (GRBs), each with an upsampling factor of $r = 2$, progressively generating 1024 and finally 2048 points.

\subsection{Performance Comparison (\textit{Answer for} Q1)}
\label{sec: exp q1} 

\begin{figure}[htbp]
\centering
\includegraphics[width=1.0\columnwidth]{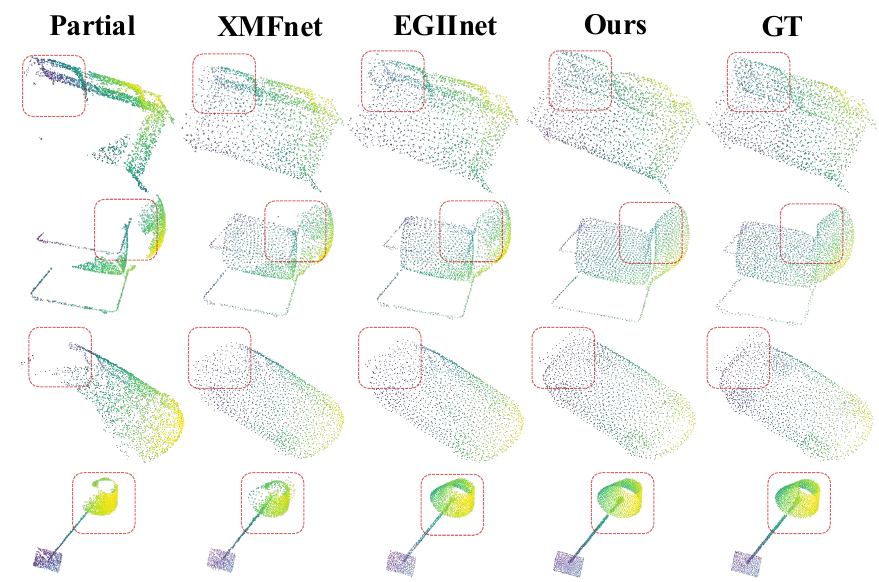} 
\caption{Qualitative comparison on ShapeNet-ViPC.}
\label{fig:qualitative_results}
\end{figure}

To answer \textbf{Q1}, we compare PGNet with state-of-the-art methods on the ShapeNet-ViPC dataset, using Chamfer Distance (CD) and F-score as evaluation metrics. As shown in Table~\ref{tab:completion_results1} and Table~\ref{tab:completion_results2}, our method establishes a new state-of-the-art across all categories, significantly outperforming the previous best method EGIInet with a 23.5\% reduction in average CD and a 7.1\% improvement in F-score. The gains are particularly pronounced in categories plagued by severe self-occlusion and challenging geometries, such as cabinet and sofa, as our paradigm robustly reconstructs large missing structures by correcting a complete prior rather than hallucinating them from sparse cues. This advantage is clearly illustrated in Figure~\ref{fig:qualitative_results}, where PGNet exhibit superior structural integrity and point uniformity.

\begin{table}[tbp]
\centering
\small 
\setlength{\tabcolsep}{3.3pt} 
\begin{tabular}{l|c|cccccccc}
\toprule
\textbf{Method} & \textbf{Avg} & \textbf{Air} & \textbf{Cab} & \textbf{Car} & \textbf{Cha} & \textbf{Lam} & \textbf{Sof} & \textbf{Tab} & \textbf{Wat} \\
\midrule
Inpaint    & 1.10         & 0.53         & 1.57         & 1.48         & 1.21         & 0.65         & 1.36         & 1.29         & 0.71         \\
PGNet  & \textbf{0.93}& \textbf{0.46}& \textbf{1.11}& \textbf{1.30}& \textbf{1.04}& \textbf{0.58}& \textbf{1.14}& \textbf{1.17}& \textbf{0.62} \\
\bottomrule
\end{tabular}
\caption{Paradigm-level comparison on ShapeNet-ViPC using CD $\times 10^{-3}$ (lower is better). “Inpaint” represents our inpainting variant.}
\label{tab:paradigm_comparison}
\end{table}
Unlike other approaches that often produce artifacts or inconsistent geometry, our method generates clean and coherent structures. Specifically, PGNet robustly recovers entire surfaces (e.g., chair seat) and delicate details (e.g., sofa pillows and armrests) by correcting a complete prior scaffold and leveraging dual-source features for refinement.

\begin{table*}[tbp]
\centering
\setlength{\tabcolsep}{3.8pt}
\begin{tabular}{c|c|cccccccc} 
\toprule 
Methods & Avg & Airplane & Cabinet & Car & Chair & Lamp & Sofa & Table & Watercraft \\ 
\midrule 
\multicolumn{10}{c}{Unimodal Methods} \\
\midrule 
FoldingNet \cite{yang2018foldingnet} & 6.271 & 5.242 & 6.958 & 5.307 & 8.823 & 6.504 & 6.368 & 7.080 & 3.882 \\
PCN \cite{yuan2018pcn} & 5.619 & 4.246 & 6.409 & 4.840 & 7.441 & 6.331 & 5.668 & 6.508 & 3.510 \\
TopNet \cite{tchapmi2019topnet} & 4.976 & 3.710 & 5.629 & 4.530 & 6.391 & 5.547 & 5.281 & 5.381 & 3.350 \\
PoinTr \cite{yu2021pointr} & 2.851 & 1.686 & 4.001 & 3.203 & 3.111 & 2.928 & 3.507 & 2.845 & 1.737 \\
SeedFormer \cite{zhou2022seedformer} & 2.902 & 1.716 & 4.049 & 3.392 & 3.151 & 3.226 & 3.603 & 2.803 & 1.679 \\
PointAttN \cite{wang2024pointattn} & 2.853 & 1.613 & 3.969 & 3.257 & 3.157 & 3.058 & 3.406 & 2.787 & 1.872 \\
\midrule 
\multicolumn{10}{c}{Multimodal Methods} \\
\midrule
ViPC \cite{zhang2021view} & 3.308 & 1.760 & 4.558 & 3.138 & 2.476 & 2.867 & 4.481 & 4.990 & 2.197 \\
CSDN \cite{zhu2023csdn} & 2.570 & 1.251 & 3.670 & 2.977 & 2.835 & 2.554 & 3.240 & 2.575 & 1.742 \\
XMFnet \cite{aiello2022cross} & 1.454 & 0.628 & 1.938 & 1.753 & 1.404 & 1.818 & 1.748 & 1.449 & 0.894 \\
EGIInet \cite{xu2024explicitly} & 1.211 & 0.552 & 1.922 & 1.659 & 1.203 & 0.777 & 1.552 & 1.227 & 0.803 \\
\midrule 
Ours & \textbf{0.926} & \textbf{0.455} & \textbf{1.111} & \textbf{1.303} & \textbf{1.038} & \textbf{0.578} & \textbf{1.139} & \textbf{1.167} & \textbf{0.615} \\
\bottomrule 
\end{tabular}
\caption{Quantitative comparison on ShapeNet-ViPC dataset using Chamfer Distance $\times 10^{-3}$ (lower is better).}
\label{tab:completion_results1}
\end{table*}

\begin{table*}[t]
\centering
\setlength{\tabcolsep}{3.8pt}
\begin{tabular}{c|c|cccccccc} 
\toprule 
Methods & Avg & Airplane & Cabinet & Car & Chair & Lamp & Sofa & Table & Watercraft \\ 
\midrule 
\multicolumn{10}{c}{Unimodal Methods} \\
\midrule 
FoldingNet \cite{yang2018foldingnet} & 0.331 & 0.432 & 0.237 & 0.300 & 0.204 & 0.360 & 0.249 & 0.351 & 0.518 \\
PCN \cite{yuan2018pcn} & 0.407 & 0.578 & 0.270 & 0.331 & 0.323 & 0.456 & 0.293 & 0.431 & 0.577 \\
TopNet \cite{tchapmi2019topnet} & 0.467 & 0.593 & 0.358 & 0.405 & 0.388 & 0.491 & 0.361 & 0.528 & 0.615 \\
PoinTr \cite{yu2021pointr} & 0.683 & 0.842 & 0.516 & 0.545 & 0.662 & 0.742 & 0.547 & 0.723 & 0.780 \\
SeedFormer \cite{zhou2022seedformer} & 0.688 & 0.835 & 0.551 & 0.544 & 0.668 & 0.777 & 0.555 & 0.716 & 0.786 \\
PointAttN \cite{wang2024pointattn} & 0.662 & 0.841 & 0.483 & 0.515 & 0.638 & 0.729 & 0.512 & 0.699 & 0.774 \\
\midrule 
\multicolumn{10}{c}{Multimodal Methods} \\
\midrule
ViPC \cite{zhang2021view} & 0.591 & 0.803 & 0.451 & 0.512 & 0.529 & 0.706 & 0.434 & 0.594 & 0.730 \\
CSDN \cite{zhu2023csdn} & 0.695 & 0.862 & 0.548 & 0.560 & 0.669 & 0.761 & 0.557 & 0.729 & 0.782 \\
XMFnet \cite{aiello2022cross} & 0.797 & 0.957 & 0.671 & 0.696 & 0.809 & 0.791 & 0.719 & 0.823 & 0.910 \\
EGIInet \cite{xu2024explicitly} & 0.836 & 0.969 & 0.691 & 0.723 & 0.847 & 0.919 & 0.756 & 0.857 & 0.927 \\
\midrule 
Ours & \textbf{0.895} & \textbf{0.985} & \textbf{0.839} & \textbf{0.804} & \textbf{0.887} & \textbf{0.954} & \textbf{0.850} & \textbf{0.881} & \textbf{0.963} \\
\bottomrule 
\end{tabular}
\caption{Quantitative comparison on ShapeNet-ViPC dataset using F-score@0.001 (higher is better).}
\label{tab:completion_results2}
\end{table*}

\begin{table}[t]
\centering
\begin{tabular}{l|l|cc}
\toprule
 \textbf{Model Variant} & \textbf{CD} $\downarrow$ & \textbf{F-score} $\uparrow$ \\
\midrule
w/o Prior Feature Grounding & 1.185 & 0.827 \\
w/o Seed Grounding & 1.219 & 0.821 \\
w/o Dual-Source Association & 1.324 & 0.803 \\
w/o Structure-Aware & 1.275 & 0.800 \\
\midrule
{\textbf{PGNet (Full Model)}} & \textbf{1.111} & \textbf{0.839} \\
\bottomrule
\end{tabular}
\caption{Ablation study of PGNet on the ShapeNet-ViPC cabinet category.}
\label{tab:ablation}
\end{table}

\subsection{Paradigm Comparison (\textit{Answer for} Q2)}
To answer \textbf{Q2}, we validate our Completion-by-Correction paradigm against a strong inpainting baseline. This baseline replaces our generative prior branch with a pretrained ResNet-18 encoder, while all other components and trainingsettings remain identical. As shown in Table 1, reverting to inpainting significantly degrades performance, increasing average CD by 18.3\% (from 0.93 to 1.10), with severe drops in occluded categories such as cabinet (+41.4\%). This underscores the limitations of synthesizing geometry directly from fused features and affirms our core premise: shifting the task from unconstrained synthesis to guided correction of a complete scaffold is a more robust strategy.

\subsection{Ablation Study (\textit{Answer for} Q3)}
To answer \textbf{Q3}, we conduct an ablation study on the cabinet category to assess the impact of key PGNet components (Table~\ref{tab:ablation}). Removing either feature-level correction (w/o Prior Feature Grounding) or scaffold alignment (w/o Seed Grounding) severely degrades performance (CD: 1.185 and 1.219), highlighting the need for an observation-consistent representation prior to refinement. Disabling dual-source association (w/o Dual-Source Association) causes the largest drop (CD: 1.324), underscoring its role in integrating high-fidelity observation geometry and structural context from the prior. Excluding the structure-aware upsampling module (w/o Structure-Aware) also harms detail recovery (CD: 1.275), confirming the value of shape-guided displacement prediction for local fidelity. In summary, these modules and their components play a pivotal role.

\section{Conclusion}
In this work, we revisited the Completion-by-Inpainting paradigm for multimodal point cloud completion, which often suffers from structural artifacts when synthesizing missing geometry from limited features. We proposed Completion-by-Correction, a more robust alternative that reframes completion as the guided refinement of a complete generative prior. Instead of direct synthesis from an incomplete fused representation, our method grounds the prior in partial observations through feature-space correction, reducing ambiguity and improving consistency. We introduced PGNet, a three-stage framework that implements this paradigm via corrective encoding, seed generation, and hierarchical refinement. Experiments on ShapeNet-ViPC show that PGNet achieves state-of-the-art accuracy and structural quality. In the future, we will to extend this approach to real-world scenarios using large-scale image-to-3D models with broader category and scene coverage.

\section*{Acknowledgments}
This work was supported in part by the Shenzhen Science and Technology Program under Grant KJZD20230923113901004, in part by the National Natural Science Foundation of China under Grants 62572501 and 62502551, and in part by the Guangzhou Yunshan Research Institute of Artificial Intelligence Security under Grant HT-99982025-0734. (Corresponding author: Di Wu.)

\bibliography{aaai2026}

\end{document}